# Multimodal Analysis of memes for sentiment extraction

Nayan Varma Alluri
Department of Computer Science and Engineering
PES University
*Bangalore, India*
nayanvarma3@gmail.com

Neeli Dheeraj Krishna
Department of Computer Science and Engineering
PES University
*Bangalore, India*
neelidheeraj2001@gmail.com

*Abstract*—Memes are one of the most ubiquitous forms of social media communication. The study and processing of memes, which are intrinsically multimedia, is a popular topic right now. The study presented in this research is based on the Memotion dataset, which involves categorising memes based on irony, comedy, motivation, and overall-sentiment. Three separate innovative transformer-based techniques have been developed, and their outcomes have been thoroughly reviewed.The best algorithm achieved a macro F1 score of 0.633 for humour classification, 0.55 for motivation classification, 0.61 for sarcasm classification, and 0.575 for overall sentiment of the meme out of all our techniques.

## I. INTRODUCTION

Due of the widespread usage and acceptance of social media, numerous new forms of communication have emerged. Memes are one of the numerous unconventional means of communication. They are multi-modal and context-sensitive communication methods. Movies, politics, TV shows, religion, and social culture are all common sources of Internet memes. Memes have two distinct characteristics: creative reproduction and intertextuality. The term "creative reproduction" refers to the blending of several scenes, whereas "intertextuality" refers to the blending of diverse cultures. Memes propagate through imitation and remixing. Mimicry refers to using the same meme in a comparable circumstance, whereas remix refers to combining and adapting the meme to the situation. Modern memes are more visually hilarious, bizarre, and diverse, resulting in a smaller percentage of people understanding their meaning and depth. It is impossible to manually examine every meme for moral and ethical policing because to the rapid growth in the volume of social media memes, as well as a lack of knowledge. As a result, an automated system that can categorise diverse memes into categories in order to separate hazardous content is urgently needed. The goal of the Memotion dataset and subsequent Memotion analysis [17] is to establish a baseline for comprehending various memes and their associated sentiment.This paper details our research into accurately predicting the comedy, sarcasm, motive, and overall attitude of various memes. A meme is multimedia in nature, consisting of both image and text. The image establishes the scene, while the text defines the situation. To construct a robust model, the paper combines both of these. Different state-of-the art techniques such as Vision transformers [2] , RoBERTa Transformers [3] and Bidirectional Long short-term memory (LSTM's) are being used.

## II. RELATED WORK

Meme sentiment analysis is a very recent field of study. Despite recent state-of-the-art publications on topics such as meme generation [4] and sentiment analysis [5], studies on sentiment analysis for memes have been limited. With the recent rise of social media, there has been a greater focus on evaluating sentiment on both image and textual data.There have been significant improvements in the fields of sentiment analysis in images and also sentiment analysis in textual data (tweets).

Detection of hate speech in tweets [6] : For the objective of detecting hate speech, the author's employed a sequence of random and glove embedding with LSTMs. The study outperforms all previous models, and the use of embedding and a gradient-boosted decision tree considerably improves the model's performance.

MELD [7] : The author's of this work discussed the MELD dataset, which is a multimodal multiparty dataset with data that includes audio, video, and photographs. For the identification of conversational emotion, a baseline model has been developed.

Expressively vulgar [8] : In this paper the authors discussed the effects of vulgarity on different sentences and the socio-cultural and pragmatic effects of the vulgarity present from tweets.

Image Sentiment Analysis Using Deep Learning [10] : In this study, the author's test the effectiveness of several approaches such as CNN, RCNN, and 3D CNNs for sentiment analysis of images.

Image sentiment analysis using deep convolutional neural networks with domain specific fine tuning [9] : Here the author uses CNN's along with domain specific tuning ie based on the type of image or description of the image.

Emotion Detection and Sentiment Analysis of Static Images [11] : The author used CNN along with face recognition models to effectively identify emotion of the images in the dataset.

## III. DATA

There are 6992 images in the memotion dataset. There is comparable value for humour, sarcasm, motivation, and overall sentiment for each image. Each sample of humour is divided into two categories: funny and not funny. Each sarcasm sample is divided into two categories: sarcastic and non-sarcastic. Each motivated sample is divided into two categories: motivational and non-motivational. Overall sentiment is divided into three categories: positive, negative, and neutral. The data set is heavily skewed (Table 1) . As a result,We employed a technique called Synthetic Minority Oversampling[12], in which the minority is resampled to equal the majority. After rebalancing the dataset according to the SMOTE technique for each emotion we have used a 80% as train data and 20% as text data.

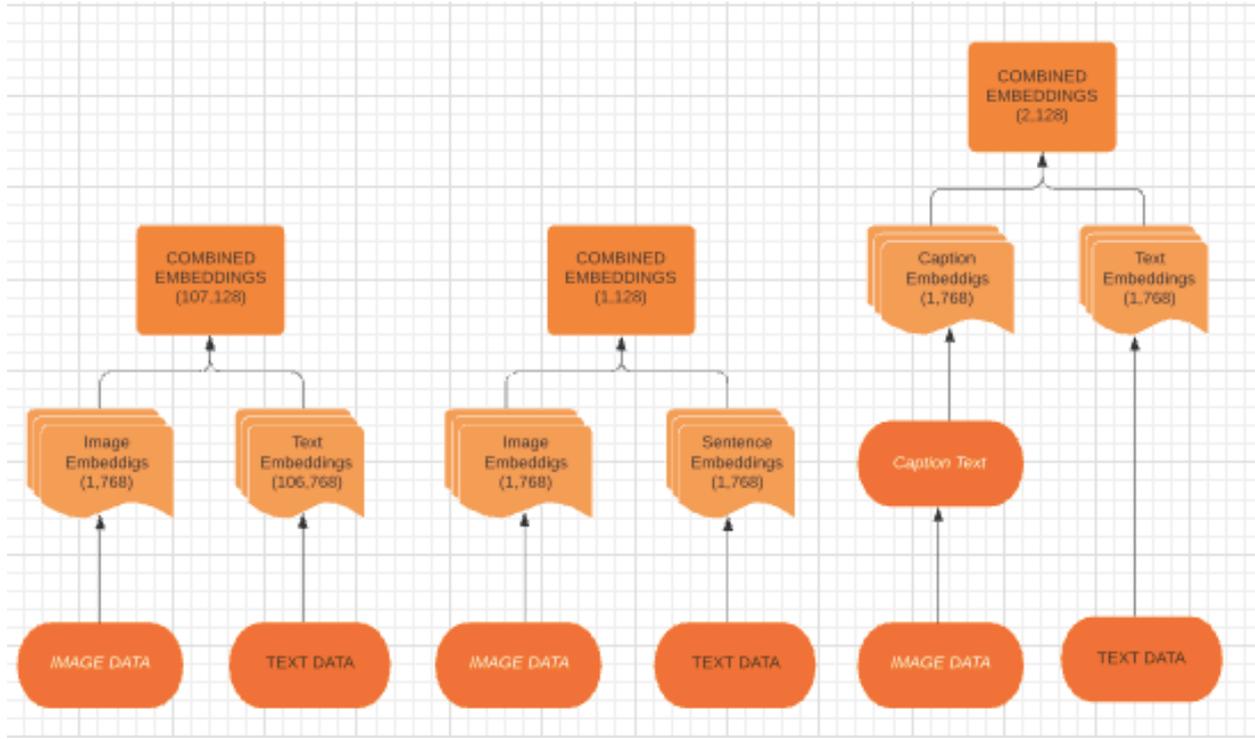

| TABLE I | | | |
|---|---|---|---|
| DISTRIBUTION OF DATA VALUES | | | |
|  | POSITIVE | NEGITIVE | NEUTRAL |
| HUMOR | 4160 | 631 | 2201 |
| SARCASM | 5341 | 1651 | 0 |
| MOTIVATION | 2467 | 4525 | 0 |
| OVERALL SENTIMENT | 1941 | 5051 | 0 |

IV. METHODOLOGY

A meme is a type of multi-media communication. It has an image on it as well as some writing carved on it. Typically, popular sources such as movies, politics, or cartoons are used to create the image. Each user's and context's text is unique. Both of them can convey the context and meaning of the situation, as well as the varied emotions. We exclusively used transformer models in this paper. To produce a strong representation of the meme, image embeddings and text embeddings are integrated. In the following subsections, we'll go over more specifics.

*A. Visual representation*

Vision Transformers [2] : Transformers are the most comprehensive representation of images currently available. They employ a sequential and gentle attention pattern that aids in the identification of the image's most influential elements. Each image is divided into 16X16 blocks, each of which is positionally encoded before being sent to the transformer network. Inside the transformer, we employ a variety of multi-head attention and normalisation techniques to ensure that the sections that have the most impact are prioritised. ViT architectures have been employed because they are more effective at representing images than normal CNN designs. Transfer learning was used to train a state-of-the-art transformer model architecture.

Image Captions : It's possible that the image won't be able to adequately represent the context. Textual features are often thought to be better at representing context than visual elements. As a result, we created a transformer-based image captioning model that can caption images and hence efficiently offer context. These captions are then put via a typical transformer model, which converts them into embeddings.

*B. Textual representation*

Word based Transformers: BERT also known as bidirectional encoder representation from transformers [13] are more effective representation of words than Fast text or glove embeddings. BERT has the advantage of being faster and being able to represent the relation between words more robustly than LSTM's or Gated Recurrent Units (GRU). They are truly directional in nature. They work on the basis of self-attention. BERT embeddings are able to accurately represent the context in the sentence. BERT training happens in 2 phases - Masked language model (MLM) and next sentence prediction(NSP). These 2 phase occur in parallel and complement each others learning process similar to a Bag of words model. In this fashion the complete sentence is encoded.

RoBERTa [3] : This model is primarily built on BERT but with significant improvements like more fine-tuning of hyper parameters , increased batch size and also a larger training database. These improvements greatly increase the robustness of this model.

Sentence based transformers [14] : This model was originally created to build cutting-edge sentence embeddings so that it could be trained with Bert's efficiency in a fraction of the time. It creates 768-dimensional embeddings and is more robust than BERT since it reduces the overhead of huge dimensionality. For higher accuracy, we employed SBERT trained on the RoBERTa model.

Bidirectional LSTM [15] : LSTMs are superior than RNNs in that they avoid long-term dependencies and therefore address the gradient descent problem. The sequence information is sent in both directions in bidirectional LSTMs, allowing the model to capture dependencies more efficiently

*C. Multimodal representation:*

Fusion method: This method is used where we simply concatenate the image and textual values along the first axis thereby creating a multidimensional data. This multidimensional data is ideal for passing to LSTM model.

Captioning model: Here we have used pretrained RESNET weights [16] and combined it with custom designed transformer model. For every image a corresponding caption is achieved.

We have designed and implemented 3 models:-

- IMGTXT Model: In this model we have used ViT model [2] and RoBERTa model [3] and their embedding have been combined using fusion method.
- IMGSEN Model: In this model we have used ViT model [2] and SBERT RoBERTa models [14] . The embeddings have been combined using fusion method
- CAPSEN Model: In this model the image is converted using captioning model and corresponding sentence is obtained. This sentence and Textual data is converted using SBERT model [14] and the embeddings are combined using fusion method.

V. EXPERIMENT

Using the techniques that were mentioned earlier we have designed various models. In this section we provide the training data along with various methods for code reproducibility.

*A. Preprocessing Data*

The textual information was unclean and some parts had no semantic meaning. The data were therefore preprocessed by the following methods:-

- All the data was converted into lowercase letters.
- All emoji that are present were converted into text using Demoji API.
- Hashtags were replaced with the word.
- The names and other user details were removed.
- Stemming was used to create more holistic data.
- Using nltk.words.words() only the words present in English language were used and the rest were removed from dataset.

*B. Training details*

All the models were trained with a batch size of 256 and Adam optimiser with learning rate 1e-3 unless specified. The data for each subcategory of emotion was balanced (Table 1) using Synthetic minority oversampling methods [12] .

- IMGTXT Model: In this model we used the concatenations of RoBERTA and ViT models and these concatenations were passed on to series of Bidirectional LSTM network and dense networks for the task of predicting 3 categories for overall sentiment and 2 categories for Sarcasm, Motivation and Humor. The model is trained for approximately 150 epochs for peak accuracy.
- IMGSEN Model: In this model we used concatenations of ViT model and SBERT - RoBERTa embeddings, and they were passed to series of bidirectional LSTM and dense layers. For Sentiment, we predict 3 categories and for the rest we predict 2 categories. The model was trained for 45 epochs for peak accuracy.
- CAPSEN Model: In this model the image is transformed into a sentence using Image captioning model and the concatenated sentence are then passed to series of Bidirectional LSTM and dense layers.For Sentiment, we predict 3 categories and for the rest we predict 2 categories. The model was trained for 75 epochs for peak accuracy. Learning rate used was 3e-4.

| TABLE II | | | |
|---|---|---|---|
| **TRAINING ACCURACIES IN PERCENTAGE** | | | |
|  | ***IMGTXT*** | ***IMGSEN*** | ***CAPSEN*** |
| HUMOR | 63.33 | 74.83 | 73.23 |
| SARCASM | 58.78 | 71.34 | 74.24 |
| MOTIVATION | 60.22 | 70.41 | 71.09 |
| OVERALL SENTIMENT | 53.60 | 73.78 | 72.15 |

| TABLE III | | | |
|---|---|---|---|
| **TESTING ACCURACIES IN PERCENTAGE** | | | |
|  | ***IMGTXT*** | ***IMGSEN*** | ***CAPSEN*** |
| HUMOR | 60.13 | 64.31 | 62.97 |
| SARCASM | 55.47 | 57.44 | 62.94 |
| MOTIVATION | 56.43 | 63.58 | 55.07 |
| OVERALL SENTIMENT | 47.87 | 60.21 | 60.25 |

| TABLE IV | | | |
|---|---|---|---|
| **F1 SCORE ON TESTING DATA** | | | |
|  | ***IMGTXT*** | ***IMGSEN*** | ***CAPSEN*** |
| HUMOR | 53.24 | 63.30 | 61.22 |
| SARCASM | 41.78 | 60.32 | 61.33 |
| MOTIVATION | 53.05 | 55.09 | 53.80 |
| OVERALL SENTIMENT | 37.12 | 57.47 | 57.79 |

VI. RESULTS AND DISCUSSION

The primary metric for training data is accuracy. We can see from Table II IMGSEN model out- performs all other models for detection of humor and overall sentiment. The CAPSEN model outperforms in detection of sarcasm and motivation.

For Testing data we have used both accuracy and F1 score as the metrics. The following observations can be made form Table III and Table IV.

- The IMGSEN model outperforms all other models in humor and motivation categories for both accuracies and F1 score.
- The CAPSEN model outperforms all other models in sarcasm and and overall sentiment.

We would want to give some comments about memes and datasets in general in order to support the results we obtained. We also studied why many cutting-edge models are unable to produce good results.

Firstly, the text in the memes cannot be semantically analysed accurately. This is because the text can be located in horizontal or vertical directions and the OCR model used can provide the findings to the intended format, where an appropriate words sequence for good semantical analysis can be provided. CASE 1 shows an example. Here the text in CASE1 is read as "everyone hand washing literally disinfection conflict avoiding any mask". This can affect the meaning of the sentence, thereby changing the sentiment.

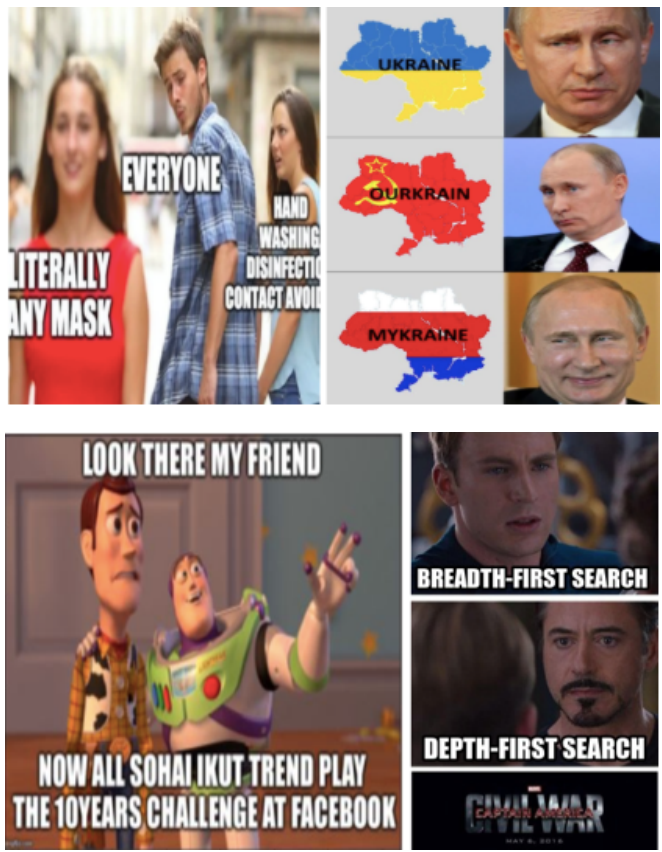

Fig.2. (A) CASE1 (B) CASE2 (C) CASE3 (4) CASE4

Second, many of the words in the image may not be found in the English language and hence have no meaning. Their significance can be deduced from their circumstances. An example is a popular Vladimir Putin meme in CASE 2. Here the meaning of MYKRAINE and OURKRAINE can be derived from the image of Putin's reaction to the memes. They also heavily depend on the current issues and its meaning may die out as time passes.

Third, modern memes are becoming increasingly ludicrous and are only understood by a small group of individuals. This is because the setting could be a real-life incident, and the attitude of the meme could shift dramatically. An example can be observed in CASE 3 where "SOHAI IKUT" can only be understood by the active players of the game.

Finally, memes have numerous levels of meaning that are borrowed from other memes, and each person's understanding of the meme in its whole may differ. This leaves a lot of room for uncertainty when it comes to categorising the meme. An example can be observed in CASE 4 where the civil war between BFS and DFS can be understood based on the movie. The user must be familiar with all the characters taking part.

The IMGTXT model's lower accuracy can be linked to the data's high dimensionality, which prevents the model from properly generalising on emotions. The average F1 score for IMGSEN model is 59.045 and average F1 score for CAPSEN model is 58.53. There is a lot of scope for improvement in CAPSEN model since the captioning model is trained on Flickr8K due to unavailability of image-captioning dataset for memes. A dedicated captioning dataset for memes can greatly improve the accuracy of CAPSEN model.

## VII. CONCLUSIONS

Overall the average the testing accuracy is 62.77% and the average F1 score is 59.045% which is significant improvement over the baseline observed but there is a large scope for improvement.

We proposed and discussed the effectiveness of using modern state-of-the-art visual and textual transformer models in enhancing accuracy and F1 score in this study. We also touched about the reasons why different models can't do a good job of analysing memes, as well as the possibilities for solving the difficulties in the future.